\documentclass[letterpaper, 10 pt, conference]{ieeeconf}  

\usepackage{amsmath,amsfonts,amssymb}
\usepackage{graphicx}
\usepackage[colorlinks=true, allcolors=blue]{hyperref}
\usepackage{multirow}
\usepackage{booktabs}
\usepackage{subfigure}
\usepackage{bbding}
\usepackage{threeparttable}

\usepackage{dsfont}
\usepackage{caption}

\usepackage{cite} 

\usepackage{amssymb}
\usepackage{amsbsy}

\usepackage[table]{xcolor}

\usepackage{booktabs} 
\usepackage{makecell}
\usepackage{multirow}
\usepackage{overpic}
\usepackage{diagbox}

\usepackage{adjustbox}
\usepackage{array}

\usepackage{verbatim}

\usepackage{color, soul}
\sethlcolor{blue}

\newcommand{\bX}{\mathbf{X}}
\newcommand{\bx}{\mathbf{x}}

\newcommand{\bZ}{\mathbf{Z}}
\newcommand{\bW}{\mathbf{W}}

\newcommand{\bL}{\mathbf{L}}
\newcommand{\bp}{\mathbf{p}}
\newcommand{\bP}{\mathbf{P}}

\IEEEoverridecommandlockouts                              

\title{\LARGE \bf
CurveFormer: 3D Lane Detection by Curve Propagation with Curve Queries and Attention
}

\author{Yifeng Bai$^{\dagger \ddagger}$, Zhirong Chen$^{\dagger}$, Zhangjie Fu, Lang Peng, Pengpeng Liang and Erkang Cheng$^{*}$ 
	\thanks{Yifeng Bai, Zhirong Chen, Zhangjie Fu, Lang Peng and Erkang Cheng are with NullMax, Shanghai, 201210, China. Pengpeng Liang is with School of Computer and Artificial Intelligence, Zhengzhou University, 450001, China.}
	\thanks{$\ddagger$ Work done during an internship at NullMax.}
	\thanks{$\dagger$ Equal contribution. *Corresponding author.}
}

\begin{document}

\maketitle
\thispagestyle{empty}
\pagestyle{empty}

\begin{abstract}
3D lane detection is an integral part of autonomous driving systems. Previous CNN and Transformer-based methods usually first generate a bird's-eye-view (BEV) feature map from the front view image, and then use a sub-network with BEV feature map as input to predict 3D lanes. Such approaches require an explicit view transformation between BEV and front view, which itself is still a challenging problem. In this paper, we propose CurveFormer, a single-stage Transformer-based method that directly calculates 3D lane parameters and can circumvent the difficult view transformation step. Specifically, we formulate 3D lane detection as a curve propagation problem by using curve queries. A 3D lane query is represented by a dynamic and ordered anchor point set. In this way, queries with curve representation in Transformer decoder iteratively refine the 3D lane detection results. Moreover, a curve cross-attention module is introduced to compute the similarities between curve queries and image features. Additionally, a context sampling module that can capture more relative image features of a curve query is provided to further boost the 3D lane detection performance. We evaluate our method for 3D lane detection on both synthetic and real-world datasets, and the experimental results show that our method achieves promising performance compared with the state-of-the-art approaches. The effectiveness of each component is validated via ablation studies as well.
\end{abstract}

\section{INTRODUCTION}
\begin{figure}[ht]
 \centering
  \begin{tabular}{c}
   \includegraphics[width=0.8\linewidth]{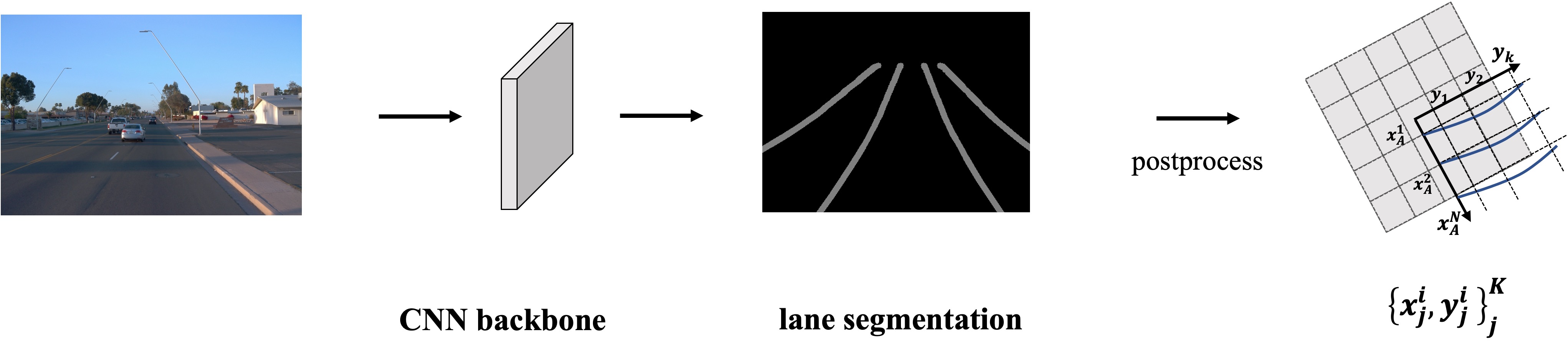}\\
   \begin{scriptsize}  
   (a) Image prediction \& post-processing. 
   \end{scriptsize} \\
   \includegraphics[width=0.8\linewidth]{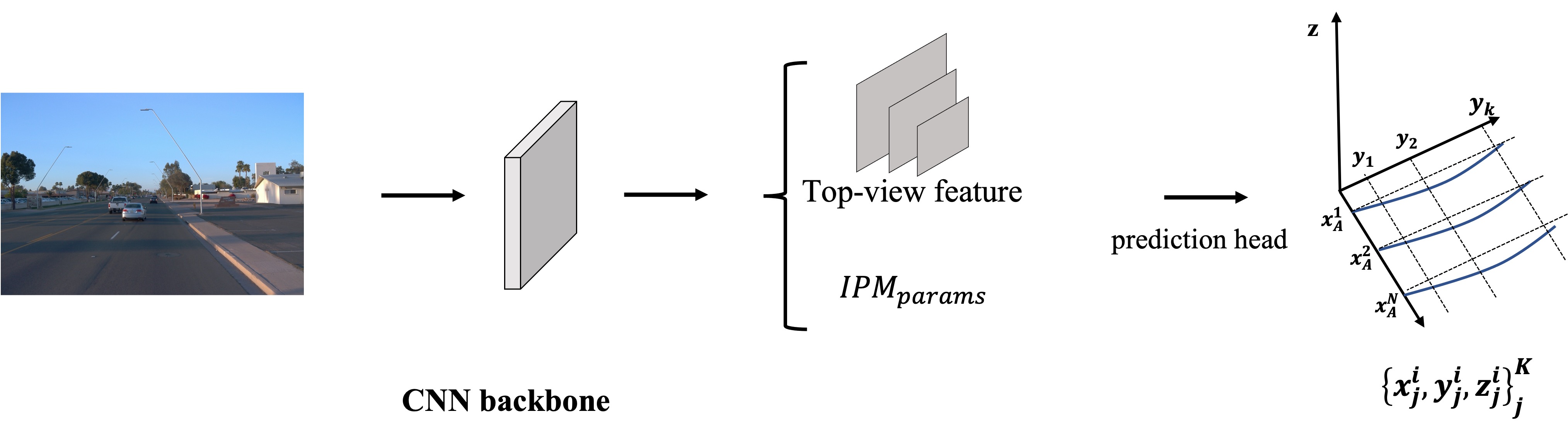}\\
   \begin{scriptsize}  
   (b) CNN-based dense BEV \& prediction.
   \end{scriptsize}  \\
   \includegraphics[width=0.8\linewidth]{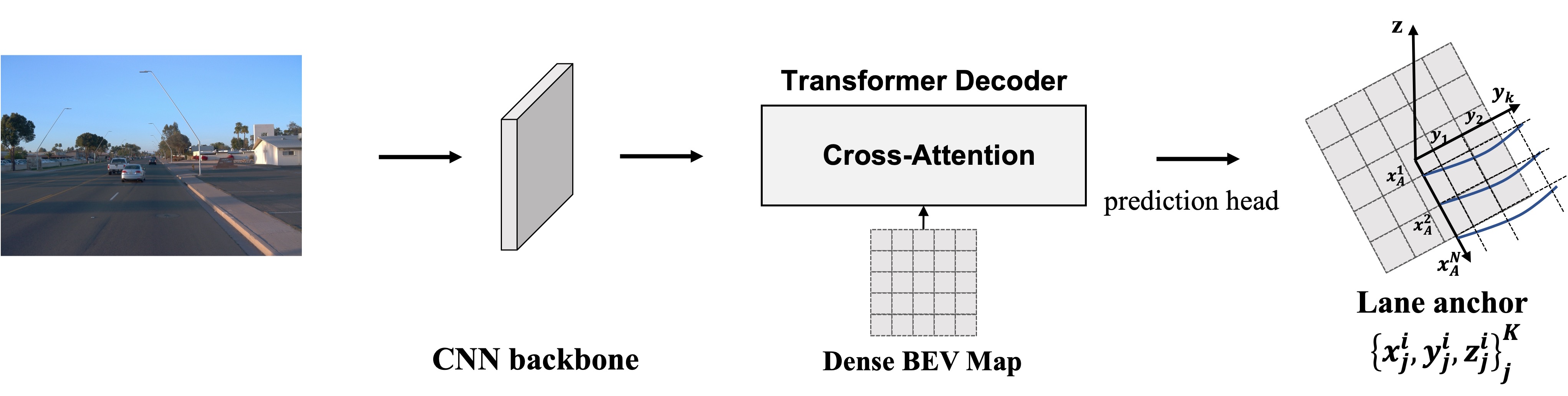}\\
   \begin{scriptsize}  
   (c) Transformer-based dense BEV \& prediction. 
   \end{scriptsize}  \\
   \includegraphics[width=0.8\linewidth]{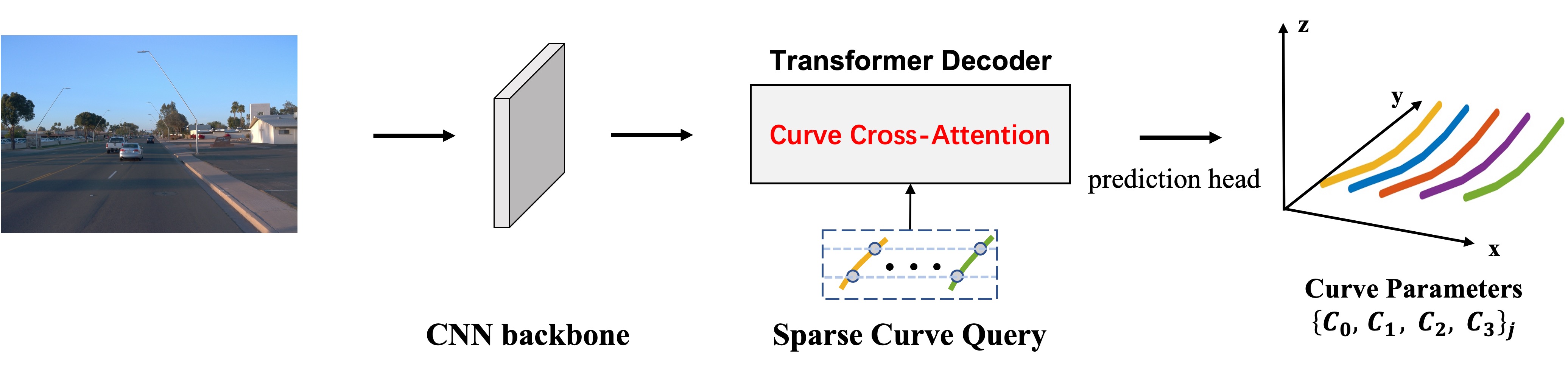}\\
   \begin{scriptsize}  
   (d) Transformer-based sparse 3D lane detection by curve query.
   \end{scriptsize}  \\
 \end{tabular}
 \caption{Comparisons of different 3D lane detection pipelines. (a) 2D image prediction and post-processing; (b) 3D lane detection with camera extrinsic prediction; (c) Transformer-based dense BEV map construction and 3D lane prediction; (d) Our proposed CurveFormer, directly provides 3D lane parameters by sparse curve queries with curve cross-attention mechanism in Transformer decoder.} 
 \label{fig_compare_methods}
\end{figure}

Lane detection is a critical component of an autonomous driving system, and it plays an important role in lane keeping assist, lane departure warning, etc. Most of the current lane detection approaches are developed with 2D images using semantic segmentation~\cite{pan2017spatial, neven2018towards, hou2019learning, zheng2020resa} or line regression~\cite{ko2020key, wang2022keypoint, chen2019pointlanenet, li2020curvelane, li2019line, 2020Keep, zheng2022clrnet, qin2020ultra}. However, downstream tasks like planning and control prefer lanes that are represented by the curve parameters in 3D space. Subject to the lack of depth information and accurate real-time camera extrinsic parameters, the projection from the image plane to the BEV perspective is prone to the error propagation problem
(as shown in Fig.~\ref{fig_compare_methods} (a)). Additionally, these methods suffer from complex and time-consuming post-processing steps, such as cluster and curve fitting. 

\begin{figure}[ht]
	\centering
	\includegraphics[width=0.6\linewidth]{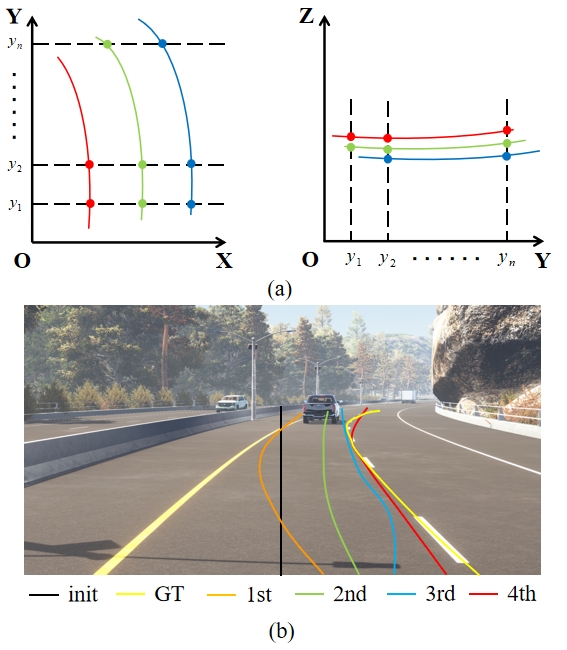}
	\caption{Illustration of the curve query representation with dynamic anchor point set (a) and iterative curve propagation the image view (b).}
	\label{fig:representation}
\end{figure}

In order to mitigate the drawbacks of post-processing in two-stage methods, CNN-based approaches have been proposed for end-to-end 3D lane detection task~\cite{garnett20193d,efrat20203d,guo2020gen,yan2022once}. As shown in Fig.~\ref{fig_compare_methods} (b), 3D-LaneNet~\cite{garnett20193d} proposes an anchor-based 3D lane representation and predicts camera pose to project 2D features with Inverse Projective Mapping (IPM). 3D-LaneNet+~\cite{efrat20203d} reformulates 3D lane as an anchor-free 
representation to consider the restriction of the lane direction, and learns lane curve clustering in the network. Gen-LaneNet~\cite{guo2020gen} proposes a virtual top view to align the BEV features projected by IPM and lanes in the real-world. Although these methods make end-to-end 3D lane detection possible, the loss of lane height and the accuracy of camera pose estimation would affect the robustness of these methods. On the contrary, ONCE~\cite{yan2022once} performs 2D lane semantic segmentation and depth estimation, and integrates these information to obtain 3D lanes. A problem of ONCE is that  depth estimation might bring about errors at the far end of the lane.

Recently, inspired by the successes of Transformer in various vision and robotic tasks~\cite{dosovitskiy2020image, carion2020end,wang2022detr3d,peng2022bevsegformer},
several Transformer-based lane detection algorithms~\cite{liu2021end,liu2022learning,chen2022persformer} have been proposed. LSTR~\cite{liu2021end} introduces Transformer to the lane detection task and predicts 2D lane parameters directly. But it encounters difficulties in representing sharp curves or lanes with complex topology. STSU~\cite{can2021structured} follows the sparse query-based framework~\cite{wang2022detr3d} to produce topologically accurate lane graphs. Similar to the above CNN-based 3D methods, CLGo~\cite{liu2022learning} applies Transformer to enhance images features and predicts 3D lanes from distance-invariant top-view image in the second stage. PersFormer~\cite{chen2022persformer} builds a dense BEV query and uses Transformer to interact queries from BEV with image features (as shown in Fig.~\ref{fig_compare_methods} (c)). Although these methods try to utilize Transformer to the 3D lane detection task, the lack of image depth or BEV map height restricts their performance as they can not obtain the features that exactly correspond to the query. 

To address the above challenges, we propose CurveFormer, a Transformer-based method for 3D lane detection (Fig.~\ref{fig_compare_methods} (d)). Lanes are defined as sparse curve queries consisting of lane confidence, two polynomials and start and end points (Fig.~\ref{fig:representation} (a)). Inspired by DAB-DETR~\cite{liu2022dab}, we propose a set of 3D dynamic anchor points to interact curve queries with image features. Since the 3D anchor point $(x,y,z)$ has height information, we can use camera extrinsic parameters to obtain accurate image features corresponding to the point. Dynamic anchor point set is iteratively refined within the sequence of Transformer decoders. We introduce a novel curve cross-attention module in the decoder part to investigate the effect of curve queries and dynamic anchor point set. Different from standard Deformable-DETR~\cite{zhu2020deformable } that directly predicts sampling offsets from the query, we introduce a context sampling unit to predict offsets from the combination of reference features and queries to guide sampling offsets learning. 
In addition, an auxiliary segmentation branch is adopted to enhance the shared CNN backbone. In this way, our design of CurveFormer lends itself to 3D lane detection.

To verify the performance of the proposed algorithm, we evaluate our CurveFormer on the Apollo Synthetic dataset~\cite{guo2020gen} and OpenLane dataset~\cite{chen2022persformer}. Our proposed CurveFormer sets a new state-of-the-art performances for 3D lane detection on the Apollo Synthetic test set. It also achieves promising performance on the OpenLane dataset compared with recently proposed Transformer-based 3D lane detection approaches. The effectiveness of each component is validated as well.

In general, our main contributions are three-fold:
\begin{itemize}
    \item We propose CurveFormer, a novel Transformer-based 3D lane detection algorithm, by formulating queries in decoder layers as dynamic anchor point set, and a curve cross-attention module is applied to compute the query-to-image similarity.
    
    \item We introduce a context sampling unit to predict offsets from the combination of reference features and queries to guide sampling offsets learning.
    
    \item Experimental results show that our method achieves promising performance compared with both CNN-based and Transformer-based state-of-the-art approaches.
\end{itemize}

\section{RELATED WORK}

\begin{figure*}[htb]
    \vspace{2mm}
	\centering
	\includegraphics[width=0.75\linewidth]{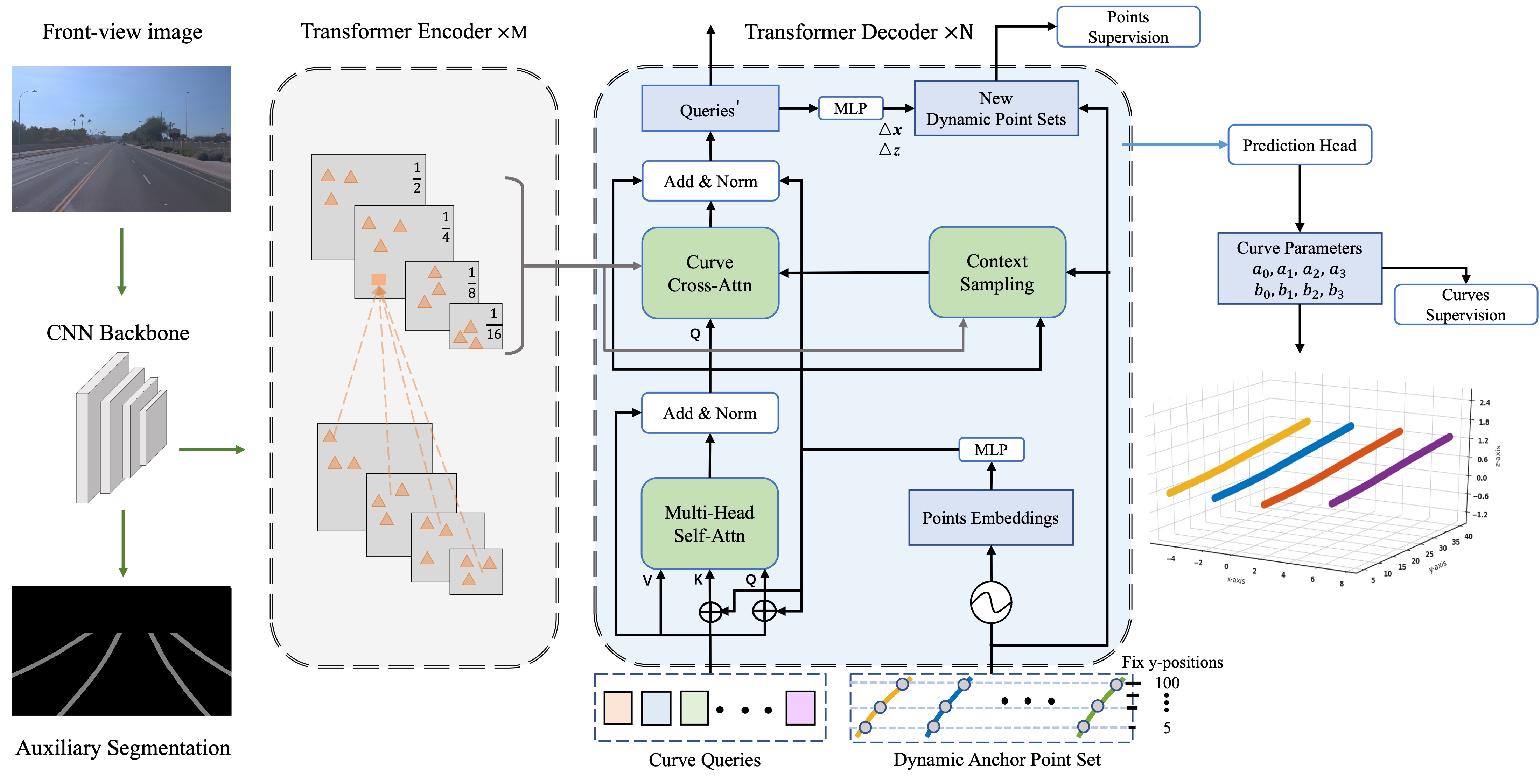}
		\caption{Overview of our proposed CurveFormer.}
	\label{fig:overview}
	\vspace{-5mm}
\end{figure*}

\noindent\textbf{2D Lane Detection.} It detects lanes in the image plane and projects them to 3D space with camera pose. In general, advanced monocular lane detectors can be categorized into segmentation approaches~\cite{pan2017spatial, neven2018towards,hou2019learning,zheng2020resa} 
and regression approaches~\cite{ko2020key, wang2022keypoint, chen2019pointlanenet,li2020curvelane,  li2019line,2020Keep,zheng2022clrnet,qin2020ultra, han2022laneformer}.

SCNN~\cite{pan2017spatial} is proposed to propagate context by slice-by-slice convolutions within feature maps. 
LaneNet~\cite{neven2018towards} introduces an instance segmentation approach for lane detection which combines a binary segmentation branch and an embedding branch. SAD~\cite{hou2019learning} allows a lane detection network to reinforce representation learning of itself without the need of additional labels and external supervisions. RESA~\cite{zheng2020resa} aggregates information in vertical and horizontal directions by shifting sliced feature maps recurrently.

Lane regression algorithms can be grouped into key points estimation~\cite{ko2020key, wang2022keypoint}, anchor-based regression~\cite{chen2019pointlanenet,li2020curvelane,  li2019line,2020Keep,zheng2022clrnet} and row-wise regression~\cite{qin2020ultra, han2022laneformer}. 
PINet~\cite{ko2020key} combines key points estimation and instance segmentation, and GANet~\cite{wang2022keypoint} represents lanes as a set of key points which are only related to the start point.
PointLaneNet~\cite{chen2019pointlanenet} and CurveLane-NAS~\cite{li2020curvelane} separate images into non-overlapping grids and regress lanes based on vertical anchors.
Line-CNN~\cite{li2019line} and LaneATT~\cite{2020Keep} regress lanes on the pre-defined ray-anchors, while CLRNet~\cite{zheng2022clrnet} dynamically refines the start point and angle of ray-anchors through pyramidal features. 
Ultra-Fast~\cite{qin2020ultra} introduces a novel row-wise classification method with remarkable speed.
Laneformer~\cite{han2022laneformer} applies row-column self-attentions to accommodate the conventional Transformer to capture the shape characteristics and semantic contexts of lanes.

Except for point regression, polynomial regression is also a method for 2D lane detection task. PolyLaneNet~\cite{tabelini2020polylanenet} uses a fully connected layer to directly predict the polynomial coefficients of lanes in the image plane. PRNet~\cite{wang2020polynomial} decomposes lane detection into three parts: polynomial regression, initial classification and height regression. Method in~\cite{van2019end} applies IPM and least square fitting to predict parabolic equations in BEV perspective. LSTR~\cite{liu2021end} introduces a Transformer-based network to predict lane parameters which reflect road structures and the camera pose.

\noindent\textbf{3D Lane Detection.} 3D lane detection has attracted more attention than its 2D counterpart recently, and a reason is that the results of the latter lack depth information and spatial transformations have the error propagation problem. 3D-LaneNet~\cite{garnett20193d} is a dual-pathway architecture based on intra-network inverse-perspective mapping and anchor-based lane representation. 3D-LaneNet+~\cite{efrat20203d} divides the BEV features into non-overlapping cells and detects lanes by regressing the lateral offset distance relative to the cell center, line angle and height offset. Method in~\cite{efrat2020semi} introduces uncertainty estimation in order to enhance the capabilities of the network. Gen-LaneNet~\cite{guo2020gen} first introduces a new geometry-guided lane anchor representation in virtual top-view coordinate frame rather than the ego-vehicle coordinate frame, and applies a specific geometric transformation to calculate 3D lane points from the network output directly. CLGo~\cite{liu2022learning} replaces the CNN backbone with Transformer to predict camera pose and polynomial parameters. PersFormer~\cite{chen2022persformer} builds a dense BEV query with known camera pose, and unifies 2D and 3D lane detection under one framework.

\section{METHOD}

\subsection{Overview}
Fig.~\ref{fig:overview} shows the overview of our CurveFormer. It consists of three major components: (1) a Shared CNN Backbone takes a single front-view image as input and outputs multi-scale feature maps; (2) a Transformer Encoder to enhance the multi-scale feature maps subsequently and (3) a curve Transformer Decoder to propagates curve queries by curve cross-attention and iteratively refine anchor point sets. Finally, a prediction head is applied to output 3D lane parameters. The $i$-th output can be represented as $\text{Pred}_i = (p_i, y_i^{start}, y_i^{end}, \{a_i, b_i\}_{r=0}^R)$,
where $p_i$ is the foreground confidence, $y_i^{start}$ and $y_i^{end}$ are start and end point in the $Y$ direction. Two polynomials of 3D lane are denoted by $a_i$ and $b_i$ with order $R$ to model a traffic lane in X-O-Y and Y-O-Z plane, respectively.

\subsection{Shared Backbone and Transformer Encoder}
The backbone takes an input image and outputs multi-scale feature maps. We add an auxiliary segmentation branch in the training stage to enhance the shared CNN backbone.

Similar to~\cite{zhu2020deformable}, in the decoder part, we apply multi-scale deformable self-attention module for each scale feature map to exchange information among different scales. The multi-scale feature maps are written as $\bX = \left\{\bx^l\right\}_{l=1}^L$.

\subsection{Representing Sparse Curve Query with Dynamic Anchor Point Set}

DAB-DETR~\cite{liu2022dab} provides a deep analysis of the role of queries for rectangle object detection which models queries as anchor boxes, i.e. 4D coordinates (x,y,w,h). Therefore, in the cross-attention module, it can leverage both the position and size information of each anchor box. Inspired by DAB-DETR, we represent queries in Transformer-based 3D lane detection with dynamic anchor point sets. As shown in Fig.~\ref{fig:representation} (a), these points are sampled at a set of fixed $Y$ locations. Typically, we denote $C_i = \{p_1=(x_1,y_1,z_1), \cdots, p_N=(x_N, y_N, z_N)\}$ as the $i$-th anchor curve.
Its corresponding content part and positional part are $Z_i \in \mathbb{R}^D$ and $P_i \in \mathbb{R}^D$, respectively. The positional query $P_i$ is calculated by:
\begin{equation}
    \text{P}_i = \text{MLP}(\text{PE}(C_i))
\label{P_i}
\end{equation}
\begin{equation}
    \text{PE}(C_i) = \text{Concat}(\text{PE}(\{x_i\}_1^N), \text{PE}(\{y_i\}_1^N), \text{PE}(\{z_i\}_1^N)),
\label{P_i}
\end{equation}
where positional encoding (PE) generates embeddings from floating numbers, and the parameters of the MLP are shared among all layers. 

\begin{figure}[ht]
 \centering
  \begin{tabular}{c}
   \includegraphics[width=0.9\linewidth]{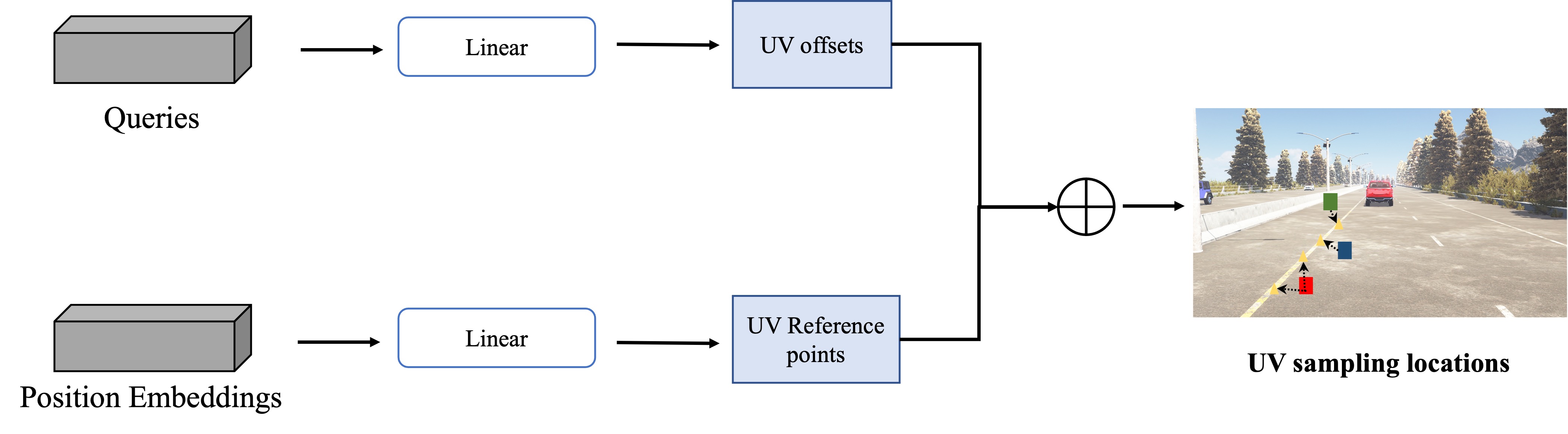}\\
   \begin{scriptsize}  
   (a)
   \end{scriptsize} \\
   \includegraphics[width=0.9\linewidth]{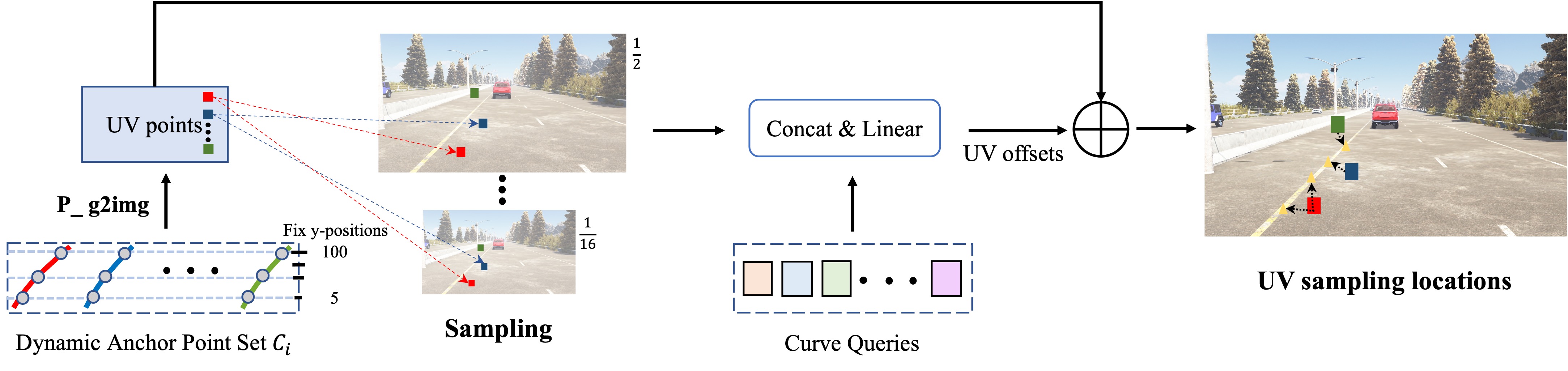}\\
   \begin{scriptsize}  
   (b)
   \end{scriptsize}  \\
 \end{tabular}
\caption{Illustration of the Context Sampling Module. (a) Deformable DETR\cite{zhu2020deformable} predicts the reference points and sampling offsets by position embedding and query separately. (b) Our context sampling module learns sampling offset by leveraging both query and image features.}
\label{fig:context-sampling}
\end{figure}

By representing a curve query as an ordered anchor point set $\{p_1 \dots p_N\}$, we can refine the curve query layer-by-layer in the Transformer decoder. Specifically, each Transformer decoder estimates relative positions $(\{\Delta x\}_1^N, \{\Delta z\}_1^N)$ by a shared parameters linear layer. In this way, the curve query representation is suitable for 3D lane detection and is able to accelerate the learning convergence via layer-by-layer refinement scheme. Fig.~\ref{fig:representation} (b) shows the iterative refine process in the image plane.

\subsection{Curve Transformer Decoder}

Our curve Transformer decoder contains a multi-head self-attention module, a context sampling module and a curve cross-attention module. 
We apply deformable attention~\cite{zhu2020deformable} in the self-attention module which focuses on a small set of key sampling points around the reference point, regardless of the spatial size of the feature map.

\noindent\textbf{Context Sampling Module} In deformable DETR~\cite{zhu2020deformable}, a learnable linear layer is used to predict offsets of the sampling locations corresponding to the reference points by queries, which are irrelevant to the image features. 
Different from it, we introduce a context sampling module to predict sampling offsets by incorporating more relative image features.
Fig.~\ref{fig:context-sampling} illustrates the difference between the standard sample offset module (a) and our context sampling module (b). 

First, a dynamic anchor point set $C_i$ is projected to the image view with camera parameters. We apply bilinear interpolation to extract features from these projected points $C_i^{2D} = \{p_1^{2D}=(u_1^i,v_1^i), \cdots, p_N^{2D}=(u_N^i, v_N^i)\}$ on multi-scale feature maps ${\bX}$. 
The final feature $f_{C_i}$ is computed by
\begin{equation}
  f_{C_i} = \frac{1}{\sum_{l=1}^{L} \sum_{n=1}^{N}\sigma_{ln} + \epsilon} \sum_{l=1}^{L} \sum_{n=1}^{N}    \bX_{l}(p_n^{2D}) \sigma_{ln}, 
\end{equation}
where $\sigma_{ln}$ is used to determine whether a projected point $p_n^{2D}$ is outside $l$-th feature map. And $\epsilon$ is a small number to avoid division by zero.

We then use a learnable linear layer to predict $K$ sampling offsets. Typically, for a curve query $\bZ_q$ with anchor point set $C_i$, the context sampling module denotes as:
\begin{align}
    \text{CS}(\{\Delta u_{nk}^i, \Delta v_{nk}^i\}) = \text{MLP} (\text{Concat}(f_{C_i}, \bZ_q)),
\label{context-sampling}
\end{align}
where $n=1,\cdots, N$ and $k=1,\cdots, K$.

\noindent\textbf{Curve Cross Attention.} We adapt deformable attention module in Deformable
DETR~\cite{zhu2020deformable} to our curve cross-attention module.
Mathematically, let $q$ be a query element in $\bZ_q$, and its anchor point set $C_i$, our curve cross-attention is calculated as:
\begin{align}
     \text{CCA}:\left(\bZ_q, C_i,\left\{\bx^l\right\}_{l=1}^L\right)= 
     \sum_{m=1}^M \bW_m \notag \\
     \left[\sum_{l=1}^L \sum_{n=1}^N A_{m l n} \cdot \bW_m^{\prime} \bx^l\left(\phi_c (p_n)+\Delta \bp_{m l  n}\right)\right],
\label{cross-attn}
\end{align}
where $(m,l,n)$ index the attention head, feature level and the sampling point. $\Delta \bp_{m l n}$ and $A_{m l n}$ denote sampling offsets and attention weights of the $n$-th sampling point in the $l$-th feature level and the $m$-th attention head. The scalar attention weight $A_{m l n}$ is normalized to sum as 1. $\phi_c (\cdot)$ re-scales the normalized coordinates to input feature maps.

\subsection{Curve Training Supervision}

\begin{table*}[t]
\centering
\begin{scriptsize}
    \caption{Comparison with previous methods on Apollo 3D Lane Synthetic Dataset. CurveFormer achieves best F-Score and AP and promising performance of X/Z error (m) on every scene set. error near and error far represents average offset within $[0m,40m]$, $[40m, 100m]$ along Y axis.}
    \label{tab:results-apollosim}
    \begin{tabular}{c|c|ccccccc}
    \toprule
    Dateset Splits & Methods & F-Score & AP & X error near & X error far & Z error near & Z error far \\
      \midrule
      \multicolumn{1}{c|}{\multirow{5}{5em}{Balaneced Scenes}}
     & 3D-LaneNet\cite{garnett20193d}   & 86.4 & 89.3 & 0.068 & 0.477 & 0.015 &  \textbf{0.202} \\
     & Gen-LaneNet\cite{guo2020gen}  & 88.1 & 90.1 & 0.061 & 0.496 & 0.012 & 0.214 \\
     & CLGo\cite{liu2022learning}         & 91.9 & 94.2 & 0.061 & 0.361 & 0.029 & 0.250 \\
     & PersFormer\cite{chen2022persformer}   & 92.9 & - &  \textbf{0.054} & 0.356 &  \textbf{0.010} & 0.234 \\
     & CurveFormer (ours) &  \textbf{95.8} &  \textbf{97.3} & 0.078 &  \textbf{0.326} & 0.018 & 0.219 \\
      \midrule
      \multicolumn{1}{c|}{\multirow{5}{5em}{Rarely Observed}} 
      & 3D-LaneNet\cite{garnett20193d}   & 72.0 & 74.6 & 0.166 & 0.855 & 0.039 & \textbf{0.521} \\
      & Gen-LaneNet\cite{guo2020gen}  & 78.0 & 79.0 & 0.139 & 0.903 & 0.030 & 0.539 \\
      & CLGo\cite{liu2022learning}         & 86.1 & 88.3 & 0.147 & \textbf{0.735} & 0.071 & 0.609 \\
      & PersFormer\cite{chen2022persformer}   & 87.5 & - & \textbf{0.107} & 0.782 & \textbf{0.024} & 0.602 \\
      & CurveFormer (ours)         & \textbf{95.6} & \textbf{97.1} & 0.182 & 0.737 & 0.039 & 0.561 &  \\
      \midrule
      \multicolumn{1}{c|}{\multirow{5}{5em}{Vivual Variants}} 
      & 3D-LaneNet\cite{garnett20193d}   & 72.5 & 74.9 & 0.115 & 0.601 & 0.032 & \textbf{0.230} \\
      & Gen-LaneNet\cite{guo2020gen}  & 85.3 & 87.2 & \textbf{0.074} & 0.538 & \textbf{0.015} & 0.232 \\
      & CLGo\cite{liu2022learning}         & 87.3 & 89.2 & 0.084 & 0.464 & 0.045 & 0.312 \\
      & PersFormer\cite{chen2022persformer}   & 89.6 & - & \textbf{0.074} & 0.430 & \textbf{0.015} & 0.266 \\
      & CurveFormer (ours)         & \textbf{90.8} & \textbf{93.0} & 0.125 & \textbf{0.410} & 0.028 & 0.254 &  \\
      \bottomrule
    \end{tabular}
\end{scriptsize}
\end{table*}

\begin{table*}[t]
  \begin{center}
  \begin{scriptsize}
    \caption{Performance comparison with other state-of-the-art 3D lane methods on OpenLane benchmark. CurveFormer outperforms previous 3D methods on five scenario sets.}
    \label{tab:results-openlane-scenario-set}
    \begin{tabular}{cccccccc}
    \toprule
      Method & All & Up\&Down & Curve & Extreme Weather & Night & Intersection & Merge\&Split \\
      \midrule
      3D-LaneNet\cite{garnett20193d} & 44.1 & 40.8 & 46.5 & 47.5 & 41.5 & 32.1 & 41.7 \\
      Gen-LaneNet\cite{guo2020gen} & 32.3 & 25.4 & 33.5 & 28.1 & 18.7 & 21.4 & 31.0 \\
      PersFormer\cite{chen2022persformer} & \textbf{50.5} & 42.4 & 55.6 & 48.6 & 46.6 & 40.0 & \textbf{50.7} \\
      CurveFormer (ours) & \textbf{50.5} & \textbf{45.2} & \textbf{56.6} & \textbf{49.7} & \textbf{49.1} & \textbf{42.9} & 45.4\\
      \bottomrule
    \end{tabular}
      \end{scriptsize}
  \end{center}
\end{table*}

\begin{table}[t]
  \begin{center}
    \caption{Comprehensive 3D Lane evaluation on OpenLane under the same metrics as Apollo 3D Synthetic. CurverFormer outperforms previous 3D methods on the metrics of near error and achieves comparable results on far error.}
    \label{tab:results-openlane-val-set}
    \begin{tiny} 
    \begin{tabular}{cccccc}
    \toprule
      Method & F-Score & X error near & X error far & Z error near & Z error far \\
      \midrule
      3D-LaneNet\cite{garnett20193d} & 44.1 & 0.479 & 0.572 & 0.367 & 0.443 \\
      Gen-LaneNet\cite{guo2020gen} & 32.3 & 0.591 & 0.684 & 0.411 & 0.521 \\
      Cond-IPM & 36.6 & 0.563 & 1.080 & 0.421 & 0.892 \\
      PersFormer\cite{chen2022persformer} & \textbf{50.5} & 0.485 &  \textbf{0.553} & 0.364 &  \textbf{0.431} \\
      CurveFormer (ours) & \textbf{50.5} & \textbf{0.340} & 0.772 &  \textbf{0.207} & 0.651 \\
      \bottomrule
    \end{tabular}
    \end{tiny} 
  \end{center}
\end{table}

In addition to the refined anchor point set $\bP = \{p_n\}_{n=1}^N$, the prediction head of our CurveFormer outputs curve parameters of $L$ 3D lanes, where $L$ is larger than the maximum number of labeled lanes across the training set. 
Similar to~\cite{liu2022learning}, we first associate the predicted curves $\text{Pred}_i = (p_i, y_i^{start}, y_i^{end}, \{a_i, b_i\}_{r=0}^R)$ and ground truth lanes $\text{GT}_i = (\hat{p}_i, \hat{y}_i^{start}, \hat{y}_i^{end}, \hat{\bL}_i=\{\hat{p}_n\}_1^N)$ by solving a bipartite matching problem, where $c \in\{0,1\}$ (0: background, 1: lane). We sample a set of 3D point $\bL_i=\{p_n\}_1^N)$ using the predicted curve parameters to compute the matching and training loss. The lane boundary (starting and ending points) is denoted by $\bL_i^b=\{ y_i^{start}, y_i^{end} \}$.

Let $\Omega = \left\{ w_l = \text{Pred}_l \right\}_{l=1}^L$ be the set of predicted 3D lanes and $\Pi = \left\{ \hat{\pi}_l = \text{GT}_l \right\}_{l=1}^L$ be the set of groundtruth.
Note that $\Pi$ is padded with non-lanes to fill enough the number of ground truth lanes to $L$. The matching problem is formulated as a cost minimization problem by searching an optimal injective function $z: \Pi \rightarrow \Omega$, where $z(l)$ is the index of a 3D lane prediction $\omega_{z(l)}$ which is assigned to $l$-th ground truth 3D lane $\hat{\pi}_l$:
\begin{equation}
    \hat{z}=\underset{z}{\arg \min } \sum_{l=1}^L D\left(\hat{\pi}_l, \omega_{z(l)}\right).
\label{matching-function}
\end{equation}
The matching cost is calculated as:
\begin{align}
    D=-\alpha_1 p_{z(l)}\left(\hat{c}_l\right)+\mathds{1}\left(\hat{c}_l=1\right) \alpha_2 \left|\hat{\bL}_l-\bL_{z(l)}\right| \notag \\
    +\mathds{1}\left(\hat{c}_l=1\right) \alpha_3 \left|\hat{\bL}_l^b-\bL_{z(l)}^{b}\right|,
\label{matching-cost}
\end{align}
where $\alpha_1$, $\alpha_2$, and $\alpha_3$ are coefficients which adjust the loss effects of classification, polynomial fitting and boundary regression, and $\mathds{1}$ is an indicator function. 

After solving Eq.~\ref{matching-function} by Hungarian algorithms~\cite{carion2020end}, the final training loss can be written as $L_{total}=L_{curve} + L_{query} + L_{seg}$, where $L_{curve}$ is the curve prediction loss, $L_{query}$ is the deep supervision of refined anchor point set for each curve, and $L_{seg}$ is an auxiliary segmentation loss.
The curve prediction loss is defined as:
\begin{align}
    L_{curve}= -\alpha_1 \log p_{\hat{z}(l)}\left(\hat{c}_l\right) + \mathds{1}\left(\hat{c}_l=1\right) \notag \\ \alpha_2\left|\hat{\bL}_l-\bL_{\hat{z}(l)}\right| 
    + \sum_{l=1}^L \mathds{1}\left(\hat{c}_l=1\right) \alpha_3 \left|\hat{\bL}_l^b-\bL_{z(l)}^{b}\right|,
\label{3D-loss}
\end{align}
where $\alpha_1$, $\alpha_2$, and $\alpha_3$ are the same coefficients with Eq.~\ref{matching-cost}, and deep supervision of refined anchor point set is:
\begin{equation}
    L_{query} = \mathds{1}\left(\hat{c}_l=1\right) \alpha_4\left|\hat{\bL}_l - \bP_{\hat{z}(l)}\right|.
\label{point-loss}
\end{equation}

\section{EXPERIMENTS}

\begin{table*}[t]
  \begin{center}
  \begin{scriptsize}
    \caption{Experimental results of CurveFormer with different number of decoder layers on Rarely Observed scenario of Apollo 3D Lane Synthetic.}
    \label{tab:decoder-layer}
    \begin{tabular}{c|c|cccccc}
    \toprule
      Dataset split & \#Layer & F-Score & AP & X error near & X error far & Z error near & Z error far \\
      \midrule
      \multicolumn{1}{c|}{\multirow{5}{5em}{Rarely Observed}}
      & 2 & 93.1 & 95.5 & 0.200 & 0.752 & 0.043 & 0.599 \\
      & 4 & \textbf{94.1} & \textbf{96.4} & \textbf{0.181} & 0.729 & \textbf{0.038} & \textbf{0.557} \\
      & 6 & 93.2 & 95.6 & 0.188 & \textbf{0.723} & 0.040 & 0.588 \\
      & 8 & 94.0 & 96.2 & 0.184 & 0.769 & 0.039 & 0.563 \\
      & 10 & 93.6 & 95.9 & 0.198 & 0.734 & \textbf{0.038} & 0.566 \\
      \bottomrule
    \end{tabular}
    \end{scriptsize}
  \end{center}
\end{table*}

\subsection{Dataset}
\textbf{Apollo 3D Lane Synthetic Dataset.} Apollo Synthetic dataset~\cite{guo2020gen} consists of over 10k 1080 × 1920 images which are built using unity 3D engine, including highway, urban, residential and downtown environments. 
The dataset is split into three different scenes: balanced scenes, rarely observed scenes and scenes with visual variations for evaluating algorithms from different perspectives.

\textbf{OpenLane Dataset.} OpenLane Dataset~\cite{chen2022persformer} is the first real world 3D lane dataset which consists of over 200K frames at a frequency of 10 FPS based on Waymo Open dataset\cite{Sun_2020_CVPR, waymo_open_dataset}. In total, it has a training set with 157k images and a validation set of 39k images. The dataset provides camera intrisics and extrinsics following the same data format as Waymo Open Dataset.

\subsection{Experiment Settings}
\textbf{Implementation Details.}  We use EfficientNet~\cite{tan2019efficientnet} as backbone which gives 4 scale feature maps. The input image is resized to size of $360 \times 480$. The 3D-space range is set to $[-30m,30m] \times [3m,103m] \times [-10m,10m]$ along $x, y$ and $z$ axis respectively. 
For curve representation, we use fixed y-positions $ \{5, 10, 15, 20, 30, 40, 50, 60, 80, 100\}$. We set coefficients to $\alpha_1=2$, $\alpha_2=5$, $\alpha_3=2$, and $\alpha_4=2$. All experiments are performed with known camera poses and intrinsic parameters provided by two datasets. Our network uses Adam optimizer~\cite{kingma2014adam}, with a base learning rate of $2 \times 10^{-4}$ and weight decay of $10^{-4}$. All models are trained from scratch with 100 epochs and the per-GPU batch size is set to 4. 

\subsection{Evaluation Metrics and Results}
\textbf{Evaluation metrics.} We follow the evaluation metrics designed by Gen-LaneNet~\cite{guo2020gen}. Point-wise Euclidean distance is calculated when a $y$-position is covered by both prediction and the ground-truth. 
For each predicted lane, we consider it matched when $ 75\% $ of its covered $y$-positions have point-wise euclidean distance less than the max-allowed distance (1.5 meters). We report Average Precision (AP) , F-score, and errors (near range and far range) to investigate the performance of our model. 

\textbf{Results on Apollo 3D Lane Synthetic Dataset.} As shown in Table.~\ref{tab:results-apollosim}, we compare our CurveFormer with CNN-based 3D lane detection and Transformer-based 3D lane detection. Experimental results verify that our method outperforms the previous state-of-the-art approaches on Apollo 3D Lane Synthetic dataset. 
CurveFormer achieves the best F-Score and AP on every scene. Compared to PersFormer~\cite{chen2022persformer} on three different scenes, CurveFormer significantly improved F-Score by $3.1\%$,  $9.0\%$ and $1.3\%$, respectively.

\textbf{Results on OpenLane Dataset.} For OpenLane dataset, we evaluate CurveFormer on entire validation set and different scenario sets.
In Table.~\ref{tab:results-openlane-scenario-set}, our CurveFormer also gets comparable results compared to previous methods on the entire validation set, and achieves the highest F-Score on five scenario sets. We present detailed comparison with previous 3D lane detection SOTAs in Table.~\ref{tab:results-openlane-val-set}.

\subsection{Ablation Study}
In this section, we analyze the effects of the proposed key components via the ablation study conducted on Apollo 3D Lane Synthetic dataset~\cite{guo2020gen}. 

\textbf{Network Output: Curve Parameters vs Anchor Point Set.} 
We compare two different network outputs for 3D lane detection, curve parameters estimation and anchor point set prediction. The latter one is further interpolated as 3D curve for evaluation. Table.~\ref{tab:represent} lists the performance comparison on Apollo 3D Lane Synthetic dataset.
It shows that curve parameter prediction largely surpasses using anchor point set as network output, due to that lane parameter prediction can preserve the geometric property of 3D lane compared to predicting separated points.

\begin{table}[t]
\centering
\begin{scriptsize}
    \caption{Results with different forms of network output on Apollo 3D Lane Synthetic Dataset. C: curve parameter estimation; P: anchor point set prediction.}
    \label{tab:represent}
    \begin{tabular}{c|c|c|cc}
    \toprule
      Dataset Splits & C & P & F-Score & AP \\
      \midrule
      \multicolumn{1}{c|}{\multirow{2}{5em}{Balaneced Scenes}} 
      & \checkmark &  & \textbf{95.8} & \textbf{97.3} \\
      &  & \checkmark & 79.1 & 80.8 \\
      \midrule
      \multicolumn{1}{c|}{\multirow{2}{5em}{Rarely Observed}} 
      & \checkmark &  & \textbf{95.6} & \textbf{97.1} \\
      &  & \checkmark & 78.2 & 80.0 \\
      \midrule
      \multicolumn{1}{c|}{\multirow{2}{5em}{Vivual Variants}} 
      & \checkmark &  & \textbf{90.8} & \textbf{93.0} \\
      &  & \checkmark & 63.8 & 64.7 \\
      \bottomrule
    \end{tabular}
\end{scriptsize}
\end{table}

\textbf{Context Sampling.} 
We study the impact of the different ways to produce sampling offsets corresponding to 3D lane reference points. As shown in Table.~\ref{tab:context-sampling}, using context sampling offset (CSO) achieves best F-Score and AP compared to using standard sampling offset (SO). The results demonstrate the significance of image-query correlation for feature aggregation in our cross-attention module.

\begin{table}[t]
\centering
\begin{scriptsize}
    \caption{Ablation study about the Context Sampling module on Apollo 3D Lane Synthetic Dataset. Baseline (Exp 1) only interacts with image features using anchor point set without sampling offsets; SO: sampling offset; CSO: context sampling offset.}
    \label{tab:context-sampling}
    \begin{tabular}{c|c|cc|cc}
    \toprule
      Dataset Splits & Exp & SO & CSO & F-Score & AP \\
      \midrule
      \multicolumn{1}{c|}{\multirow{3}{5em}{Balaneced Scenes}} 
      & 1 &  &  & 95.6 & 97.2 \\
      & 2 & \checkmark &  & 95.4 & 97.2 \\
      & 3 &  & \checkmark  & \textbf{95.8} & \textbf{97.3} \\
      \midrule
      \multicolumn{1}{c|}{\multirow{3}{5em}{Rarely Observed}} 
      & 1 &  &  & 94.0 & 96.3 \\
      & 2 & \checkmark &  & 95.4 & 97.0 \\
      & 3 &  & \checkmark  & \textbf{95.6} & \textbf{97.1} \\
      \midrule
      \multicolumn{1}{c|}{\multirow{3}{5em}{Vivual Variants}} 
      & 1 &  &  & 87.1 & 89.2 \\
      & 2 & \checkmark &  & 88.2 & 90.1  \\
      & 3 &  & \checkmark  & \textbf{90.8} & \textbf{93.0} \\
      \bottomrule
    \end{tabular}
\end{scriptsize}
\vspace{-2mm}
\end{table}

\textbf{Number of Decoder Layer.} 
We vary the number of decoder layers and the performance of the model is shown in Table.~\ref{tab:decoder-layer}. It shows that using 4 decoder layers in our network achieves best performance. We can use few decoder layers due to curve propagation scheme of our CurveFormer method. Therefore, we set the number of decoder layers in our CurveFormer to 4 by default in the experiments.

\textbf{Auxiliary Segmentation.} Lastly, we study the effect of the auxiliary segmentation branch. 
Experimental results show that the auxiliary segmentation branch can slightly improve F-Score by 0.13, AP by 0.06 on the Balanced Scenes test set.

\section{CONCLUSIONS}

In this paper, we introduce CurveFormer, a Transformer-based 3D lane detection method. It uses dynamic anchor point set to construct queries, and refines it layer-by-layer in Transformer decoders. In addition, to attend to more relevant image features, we present a curve cross-attention module and a context sampling module to compute the key-to-image similarity. 
In the experiments, we show that CurveFormer achieves promising results compared with both CNN-based and Transformer-based approaches. In future work, we would like to explore video-based 3D lane detection for autonomous driving.













\bibliographystyle{IEEEtran}
\bibliography{refs}

\begin{thebibliography}{10}
\providecommand{\url}[1]{#1}
\csname url@rmstyle\endcsname
\providecommand{\newblock}{\relax}
\providecommand{\bibinfo}[2]{#2}
\providecommand\BIBentrySTDinterwordspacing{\spaceskip=0pt\relax}
\providecommand\BIBentryALTinterwordstretchfactor{4}
\providecommand\BIBentryALTinterwordspacing{\spaceskip=\fontdimen2\font plus
\BIBentryALTinterwordstretchfactor\fontdimen3\font minus
  \fontdimen4\font\relax}
\providecommand\BIBforeignlanguage[2]{{%
\expandafter\ifx\csname l@#1\endcsname\relax
\typeout{** WARNING: IEEEtran.bst: No hyphenation pattern has been}%
\typeout{** loaded for the language `#1'. Using the pattern for}%
\typeout{** the default language instead.}%
\else
\language=\csname l@#1\endcsname
\fi
#2}}

\bibitem{pan2017spatial}
X.~Pan, J.~Shi, P.~Luo, X.~Wang, and X.~Tang, ``Spatial as deep: Spatial cnn
  for traffic scene understanding,'' \emph{arXiv preprint arXiv:1712.06080},
  2017.

\bibitem{neven2018towards}
D.~Neven, B.~De~Brabandere, S.~Georgoulis, M.~Proesmans, and L.~Van~Gool,
  ``Towards end-to-end lane detection: an instance segmentation approach,'' in
  \emph{2018 IEEE intelligent vehicles symposium (IV)}.\hskip 1em plus 0.5em
  minus 0.4em\relax IEEE, 2018, pp. 286--291.

\bibitem{hou2019learning}
Y.~Hou, Z.~Ma, C.~Liu, and C.~C. Loy, ``Learning lightweight lane detection
  cnns by self attention distillation,'' in \emph{Proceedings of the IEEE/CVF
  International Conference on Computer Vision}, 2019, pp. 1013--1021.

\bibitem{zheng2020resa}
T.~Zheng, H.~Fang, Y.~Zhang, W.~Tang, Z.~Yang, H.~Liu, and D.~Cai, ``Resa:
  Recurrent feature-shift aggregator for lane detection,'' \emph{arXiv preprint
  arXiv:2008.13719}, 2020.

\bibitem{ko2020key}
Y.~Ko, J.~Jun, D.~Ko, and M.~Jeon, ``Key points estimation and point instance
  segmentation approach for lane detection,'' \emph{arXiv preprint
  arXiv:2002.06604}, 2020.

\bibitem{wang2022keypoint}
J.~Wang, Y.~Ma, S.~Huang, T.~Hui, F.~Wang, C.~Qian, and T.~Zhang, ``A
  keypoint-based global association network for lane detection,'' in
  \emph{Proceedings of the IEEE/CVF Conference on Computer Vision and Pattern
  Recognition}, 2022, pp. 1392--1401.

\bibitem{chen2019pointlanenet}
Z.~Chen, Q.~Liu, and C.~Lian, ``Pointlanenet: Efficient end-to-end cnns for
  accurate real-time lane detection,'' in \emph{2019 IEEE Intelligent Vehicles
  Symposium (IV)}.\hskip 1em plus 0.5em minus 0.4em\relax IEEE, 2019, pp.
  2563--2568.

\bibitem{li2020curvelane}
Z.~Li, ``Curvelane-nas: Unifying lane-sensitive architecture search and
  adaptive point blending,'' 2020.

\bibitem{li2019line}
X.~Li, J.~Li, X.~Hu, and J.~Yang, ``Line-cnn: End-to-end traffic line detection
  with line proposal unit,'' \emph{IEEE Transactions on Intelligent
  Transportation Systems}, vol.~21, no.~1, pp. 248--258, 2019.

\bibitem{2020Keep}
L.~Tabelini, R.~Berriel, T.~M. Paixo, C.~Badue, A.~D. Souza, and
  T.~Oliveira-Santos, ``Keep your eyes on the lane: Real-time attention-guided
  lane detection,'' 2020.

\bibitem{zheng2022clrnet}
T.~Zheng, Y.~Huang, Y.~Liu, W.~Tang, Z.~Yang, D.~Cai, and X.~He, ``Clrnet:
  Cross layer refinement network for lane detection,'' in \emph{Proceedings of
  the IEEE/CVF Conference on Computer Vision and Pattern Recognition}, 2022,
  pp. 898--907.

\bibitem{qin2020ultra}
Z.~Qin, H.~Wang, and X.~Li, ``Ultra fast structure-aware deep lane detection,''
  \emph{arXiv preprint arXiv:2004.11757}, 2020.

\bibitem{garnett20193d}
N.~Garnett, R.~Cohen, T.~Pe'er, R.~Lahav, and D.~Levi, ``3d-lanenet: end-to-end
  3d multiple lane detection,'' in \emph{Proceedings of the IEEE/CVF
  International Conference on Computer Vision}, 2019, pp. 2921--2930.

\bibitem{efrat20203d}
N.~Efrat, M.~Bluvstein, S.~Oron, D.~Levi, N.~Garnett, and B.~E. Shlomo,
  ``3d-lanenet+: Anchor free lane detection using a semi-local
  representation,'' \emph{arXiv preprint arXiv:2011.01535}, 2020.

\bibitem{guo2020gen}
Y.~Guo, G.~Chen, P.~Zhao, W.~Zhang, J.~Miao, J.~Wang, and T.~E. Choe,
  ``Gen-lanenet: A generalized and scalable approach for 3d lane detection,''
  in \emph{European Conference on Computer Vision}.\hskip 1em plus 0.5em minus
  0.4em\relax Springer, 2020, pp. 666--681.

\bibitem{yan2022once}
F.~Yan, M.~Nie, X.~Cai, J.~Han, H.~Xu, Z.~Yang, C.~Ye, Y.~Fu, M.~B. Mi, and
  L.~Zhang, ``Once-3dlanes: Building monocular 3d lane detection,'' in
  \emph{Proceedings of the IEEE/CVF Conference on Computer Vision and Pattern
  Recognition}, 2022, pp. 17\,143--17\,152.

\bibitem{dosovitskiy2020image}
A.~Dosovitskiy, L.~Beyer, A.~Kolesnikov, D.~Weissenborn, X.~Zhai,
  T.~Unterthiner, M.~Dehghani, M.~Minderer, G.~Heigold, S.~Gelly,
  \emph{et~al.}, ``An image is worth 16x16 words: Transformers for image
  recognition at scale,'' \emph{arXiv preprint arXiv:2010.11929}, 2020.

\bibitem{carion2020end}
N.~Carion, F.~Massa, G.~Synnaeve, N.~Usunier, A.~Kirillov, and S.~Zagoruyko,
  ``End-to-end object detection with transformers,'' in \emph{European
  conference on computer vision}.\hskip 1em plus 0.5em minus 0.4em\relax
  Springer, 2020, pp. 213--229.

\bibitem{wang2022detr3d}
Y.~Wang, V.~C. Guizilini, T.~Zhang, Y.~Wang, H.~Zhao, and J.~Solomon, ``Detr3d:
  3d object detection from multi-view images via 3d-to-2d queries,'' in
  \emph{Conference on Robot Learning}.\hskip 1em plus 0.5em minus 0.4em\relax
  PMLR, 2022, pp. 180--191.

\bibitem{peng2022bevsegformer}
L.~Peng, Z.~Chen, Z.~Fu, P.~Liang, and E.~Cheng, ``Bevsegformer: Bird's eye
  view semantic segmentation from arbitrary camera rigs,'' \emph{arXiv preprint
  arXiv:2203.04050}, 2022.

\bibitem{liu2021end}
R.~Liu, Z.~Yuan, T.~Liu, and Z.~Xiong, ``End-to-end lane shape prediction with
  transformers,'' in \emph{Proceedings of the IEEE/CVF winter conference on
  applications of computer vision}, 2021, pp. 3694--3702.

\bibitem{liu2022learning}
R.~Liu, D.~Chen, T.~Liu, Z.~Xiong, and Z.~Yuan, ``Learning to predict 3d lane
  shape and camera pose from a single image via geometry constraints,'' in
  \emph{Proceedings of the AAAI Conference on Artificial Intelligence},
  vol.~36, no.~2, 2022, pp. 1765--1772.

\bibitem{chen2022persformer}
L.~Chen, C.~Sima, Y.~Li, Z.~Zheng, J.~Xu, X.~Geng, H.~Li, C.~He, J.~Shi,
  Y.~Qiao, and J.~Yan, ``Persformer: 3d lane detection via perspective
  transformer and the openlane benchmark,'' in \emph{European Conference on
  Computer Vision (ECCV)}, 2022.

\bibitem{can2021structured}
Y.~B. Can, A.~Liniger, D.~P. Paudel, and L.~Van~Gool, ``Structured
  bird's-eye-view traffic scene understanding from onboard images,'' in
  \emph{Proceedings of the IEEE/CVF International Conference on Computer
  Vision}, 2021, pp. 15\,661--15\,670.

\bibitem{liu2022dab}
S.~Liu, F.~Li, H.~Zhang, X.~Yang, X.~Qi, H.~Su, J.~Zhu, and L.~Zhang,
  ``Dab-detr: Dynamic anchor boxes are better queries for detr,'' \emph{arXiv
  preprint arXiv:2201.12329}, 2022.

\bibitem{zhu2020deformable}
X.~Zhu, W.~Su, L.~Lu, B.~Li, X.~Wang, and J.~Dai, ``Deformable detr: Deformable
  transformers for end-to-end object detection,'' \emph{arXiv preprint
  arXiv:2010.04159}, 2020.

\bibitem{han2022laneformer}
J.~Han, X.~Deng, X.~Cai, Z.~Yang, H.~Xu, C.~Xu, and X.~Liang, ``Laneformer:
  Object-aware row-column transformers for lane detection,'' \emph{arXiv
  preprint arXiv:2203.09830}, 2022.

\bibitem{tabelini2020polylanenet}
L.~Tabelini, R.~Berriel, T.~M. Paix{\~a}o, C.~Badue, A.~F. De~Souza, and
  T.~Oliveira-Santos, ``Polylanenet: Lane estimation via deep polynomial
  regression,'' \emph{arXiv preprint arXiv:2004.10924}, 2020.

\bibitem{wang2020polynomial}
B.~Wang, Z.~Wang, and Y.~Zhang, ``Polynomial regression network for
  variable-number lane detection,'' in \emph{European Conference on Computer
  Vision}.\hskip 1em plus 0.5em minus 0.4em\relax Springer, 2020, pp. 719--734.

\bibitem{van2019end}
W.~Van~Gansbeke, B.~De~Brabandere, D.~Neven, M.~Proesmans, and L.~Van~Gool,
  ``End-to-end lane detection through differentiable least-squares fitting,''
  in \emph{Proceedings of the IEEE International Conference on Computer Vision
  Workshops}, 2019, pp. 0--0.

\bibitem{efrat2020semi}
N.~Efrat, M.~Bluvstein, N.~Garnett, D.~Levi, S.~Oron, and B.~E. Shlomo,
  ``Semi-local 3d lane detection and uncertainty estimation,'' \emph{arXiv
  preprint arXiv:2003.05257}, 2020.

\bibitem{Sun_2020_CVPR}
P.~Sun, H.~Kretzschmar, X.~Dotiwalla, A.~Chouard, V.~Patnaik, P.~Tsui, J.~Guo,
  Y.~Zhou, Y.~Chai, B.~Caine, V.~Vasudevan, W.~Han, J.~Ngiam, H.~Zhao,
  A.~Timofeev, S.~Ettinger, M.~Krivokon, A.~Gao, A.~Joshi, Y.~Zhang, J.~Shlens,
  Z.~Chen, and D.~Anguelov, ``Scalability in perception for autonomous driving:
  Waymo open dataset,'' in \emph{Proceedings of the IEEE/CVF Conference on
  Computer Vision and Pattern Recognition (CVPR)}, June 2020.

\bibitem{waymo_open_dataset}
``Waymo open dataset: An autonomous driving dataset,'' 2019.

\bibitem{tan2019efficientnet}
M.~Tan and Q.~Le, ``Efficientnet: Rethinking model scaling for convolutional
  neural networks,'' in \emph{International conference on machine
  learning}.\hskip 1em plus 0.5em minus 0.4em\relax PMLR, 2019, pp. 6105--6114.

\bibitem{kingma2014adam}
D.~P. Kingma and J.~Ba, ``Adam: A method for stochastic optimization,''
  \emph{arXiv preprint arXiv:1412.6980}, 2014.

\end{thebibliography}

\end{document}